\documentclass[a4paper]{article}

\usepackage{amsmath}
\usepackage{INTERSPEECH2021, url, multirow, amsmath, pifont, amssymb, makecell}
\usepackage{algorithmic}
\usepackage{algorithm}
\usepackage{lineno,hyperref}
\usepackage[utf8x]{inputenc} 
\usepackage{graphicx}
\usepackage{babel,blindtext}
\usepackage{lipsum,graphicx,subcaption}

\DeclareUnicodeCharacter{2212}{\minus}

\title{Attacker Behaviour Profiling using Stochastic Ensemble of Hidden Markov Models}
\name{Soham Deshmukh, Rahul Rade, Faruk Kazi}
\address{{\fontsize{10}{12}\selectfont Centre of Excellence in Complex and Nonlinear Dynamical Systems, VJTI}}
\email{\{ssdeshmukh\_b15, rsrade\_b15, fskazi\}@el.vjti.ac.in}

\begin{document}

\maketitle
\begin{abstract}
Cyber threat intelligence is one of the emerging areas of focus in information security. Much of the recent work has focused on rule-based methods and detection of network attacks using Intrusion Detection algorithms. In this paper, we propose a framework for inspecting and modelling the behavioural aspect of an attacker to obtain better insight predictive power on his future actions. For modelling, we propose a novel semi-supervised algorithm called \textit{Fusion Hidden Markov Model (FHMM)} which is more robust to noise, requires comparatively less training time, and utilizes the benefits of ensemble learning to better model temporal relationships in data. This paper evaluates the performance of FHMM and compares it with both traditional algorithms like Markov Chain, Hidden Markov Model (HMM) and recently developed Deep Recurrent Neural Network (Deep RNN) architectures. We conduct the experiments on a dataset consisting of real data attacks on a Cowrie honeypot system. FHMM provides accuracy comparable to deep RNN architectures at significant lower training time. Given these experimental results, we recommend using FHMM for modelling discrete temporal data for significantly faster training and better performance than existing methods.
\end{abstract}

\section{Introduction}

The existing cyber security tools focus on reactive methods and algorithms as a major part of their cyber security arsenal. In current world where organizations are highly digital, a single vulnerability can lead to penetrative attack negatively affecting business on a large scale. Moreover, attackers now are leveraging automation and cloud to scale their attacks faster and infiltrate systems in record break time. Therefore, it is advisable for organization to stay one step ahead of attackers and be able quickly foresee where and when they will strike. Knowing the potential strike points or actions of attacker, the organization can take necessary steps for mitigating cyber risks to organization’s business. 

The attacker may use multiple access point to deploy his attack. Some attackers use persistent attack strategy consisting of a sequence of attack behaviours continuously until the intended target system is compromised \cite{razzaq}. Some attackers tend to stay in the system for long time do trivial task to bypass IDS and other things then later perform malicious act. These type of attacks are difficult to track and detect. Moreover, insider threats have become more prevalent and the breach level index shows that almost 40\% of the breaches are due to poor employee awareness of cyber security \cite{gemalto}. Although raw data in the form of logs is abundantly available, it is difficult and time-consuming to extract meaningful information based on which proactive measures can be employed. Thus, it has become indispensable to develop a solution which provides threat intelligence capabilities to combat these attacks by being both proactive and responsive.

In order to identify and mitigate security breaches, three major types of network analysis are done: signature based, anomaly based, and hybrid. Signature based techniques try to detect attacks based on their signature \cite{modi}. They have less false positive cases but are useless against new (zero-day) attacks or attack not present in their database. Anomaly based techniques \cite{mukherjee} model normal behaviour and detect deviations from the normal behaviour. They have the capacity to detect zero-day attacks. Additionally, they can be used to generate dataset for signature based methods. However, the user or employee can deviate from standard pattern of operation and this leads to high amount of false alarms generated by anomaly based approaches. Hybrid methods combine both signature and anomaly detection to increase detection \cite{viegas} of threats and reduce false positive alarms. Most of the Machine learning and Deep learning based methods are hybrid methods. However, all of these methods are supervised learning methods where the algorithms need to be provided with labelled data in the form of database, or time-series to accurately make predictions. 

There is a need of unsupervised or semi-supervised algorithm which can work on raw data dumps to provide intimation of a potential attack or threat beforehand. For the algorithm to be effective and employable for predicting threats at organization level - 

\begin{itemize}
    \item \textit{Modelling capacity}: The algorithm must have sufficient modelling capacity for detecting nonlinear patterns in sequential data.
    \item \textit{Time constraint}: The algorithm must be scalable and parallelizable to train on large data the organization generates. It should provide predictions with low latency and have low training time as the attackers can utilize the downtime of algorithm to its advantage.
    \item \textit{Data Imbalance constraint}: The algorithm should be able to handle imbalanced data distribution where threat or breaches are needle in a hack stack of ‘normal’ behaviour user or employee behaviour pattern.
    \item \textit{False Positives}: The algorithm should be able to incorporate uncertainty and possibly inform uncertain predictions rather than generating false positives.
\end{itemize}

With view to satisfy these constraints, we propose a new algorithm called \textit{Fusion Hidden Markov Model} which exploits the benefits of ensemble learning \cite{dietterich} to produce better results. We train a set of diverse HMMs on $K$ different low-correlated partitions of data and amalgamate the predictions of these models using a nonlinear weight function. The nonlinear function uses the posterior distribution over these HMMs to generate a single output. One advantage of fitting a set of HMMs is that each model learns temporal features unique to a group of similar attack patterns, thus making FHMM less susceptible to noise. We employ this approach to effectively model attack patterns on file system based honeypots. The predictive power of this model can be used to comprehend the mindset of the attacker beforehand and would help in preventing systems from being compromised.

The remainder of the paper is structured as follows. Recent work in the field of cyber security has been elaborated in Section~\ref{sect:related-work}. Section~\ref{sect:research-context} provides a detailed problem formulation along with an overview of the proposed system architecture. In Section~\ref{sect:methodology},  we explain the proposed  FHMM algorithm. Section~\ref{sect:exp} includes experiments performed. We analyze the results of these experiments and provide a comprehensive comparison of FHMM with other commonly used sequence  models for modelling attacker behaviour. Lastly, Section~\ref{sect:conclusion} summarizes and concludes the paper.

\section{Related Work}
\label{sect:related-work}
Threat intelligence or predictive threat intelligence is a newly coined term which focuses on predictive methods for cyber security. \cite{cardenas} suggested the use of big data analytics for analyzing and handling the huge amount of data traffic. One of the most prominently used machine learning based approach for insider threat or breach detection is user behaviour profiling \cite{alqurishi}. In this approach, the sequential actions of employee or student are modelled to form a profile for that specific user. The most common patterns of actions are termed as normal behaviour and any deviation from this predefined path is considered as a deviation. This anomalous activity is then marked for further investigation or potential attack. For modelling ‘normal’ behaviour of employees’ variety of strategies are used. For detecting long-term anomalies in cloud data, \cite{vallis} proposed using Extreme Studentized Deviate test (ESD) and employing time series decomposition and robust statistics for detecting anomalies. While other methods in literature rely on combining user action data with behavioural and personality characteristics, \cite{adaboost-ids, haque} propose the use of bagging and boosting algorithms for intrusion detection \cite{denning}. 
Apart from using collected or logged data on attacker actions, past research literature has shown that behavioural and personality characteristics can provide great insight in profiling the attack. This particular example is shown by \cite{brdiczka}, which uses  structural anomaly detection combined with psychological profiling, in order to reduce false positives compared to traditional anomaly detection. Work by \cite{spitzner} explores how honeypot technology can be used to detect, identify and gather information on these specific threats. Ultimately the task is to find out if there is an activity being planned, and if so, find what stage the planning is in. \cite{granstorm} showed how Bernoulli filter can be used to detect the presence of HMM in structured action data along with minimum complexity that an HMM would need to involve in order that it be detectable with reasonable fidelity. 

Recently, a lot of work has focused on rigorous analysis of data collected by honeypots using machine learning and statistical techniques. Despite potential limitations on the availability of data in the majority of prior research, intrinsic patterns in cyber attacks have been identified. For example, \cite{zhan} demonstrated the presence of long-range temporal dependencies in attack patterns captured by low-interaction honeypots. Owning to the presence of long-range dependencies in the rich information captured by honeypots, various frameworks have been proposed which exploit these statistical properties for modelling cyber attacks \cite{kaaniche}. Other frameworks proposed involve \cite{thonnard} which deals with graph-based clustering of time-series attack patterns thereby identifying the activities of worms and botnets present in the honeypot traffic. A technique proposed in \cite{almotairi} employs a similarity metric called squared prediction error (SPE) for computing distance between observation projected in residual space and the $k$-dimensional hyperspace determined by principal component analysis (PCA). This metric was used to identify new attack patterns in the honeypot logs. On the other hand, the intent behind cyber attacks also provide a fair indication of the target of the attacker and several approaches for intent prediction using HMMs \cite{zhang} have been developed. To detect intrusions, \cite{ye} employed two EWMA techniques to detect anomalous changes in event intensity specifically for correlated and uncorrelated data. Behaviour-rule based intrusion detection methods where also investigated in domain of cyber physical systems particularly in Smart Grid Applications by \cite{mitchell} which demonstrated that detecting attackers based on behaviour features led to low false positives. On the other hand, \cite{shi} used stochastic modelling framework particularly using finite state Hidden Markov Model to solve joint state and attack estimation problem. \cite{jonsson} found time feature to be extremely important in modelling attacker behaviour and the attack process can be split into three phases namely the learning phase, standard phase and innovative attack phase. \cite{qdla} suggested the use of game theory approach by formulating a Bayesian game to understand the attacker-defender interactions in honeypot-enabled networks. 

The use of Hidden Markov Models for modelling normal behaviour (against attacker actions) was proposed by \cite{rashid} for detection insider threat. There is significant work been done in improving HMM and its learning algorithm. The ability of HMM has to model sequence is dependent on the structure of HMM. Though the integration over all possible model structures are not possible, structures suited to specific domain like profile HMM \cite{krogh} are developed. In the context of biological sequence analysis, researchers have used genetic algorithm to determine the structure of HMM \cite{won}. \cite{johansson} employs a unified Bayesian treatment to derive posterior probability for different model structures within class of multinomial,  Markov, Hidden Markov models without assuming prior knowledge of transition probabilities. The searching of HMM structure followed by learning makes the approach infeasible in domains of cyber security where latency is of utmost importance.

\section{Preliminary Research Context}
\label{sect:research-context}

\subsection{Problem Formulation}

Given previous actions ${S_i}|t$ of attacker at each timestep $t$, we intend to predict the next action attacker will likely take. Here ${S_i}$ are discrete states corresponding to attacker action where $i~\in~(0, N)$.

The goal of our system is to achieve accurate attack prediction with low prediction latency and training time. The algorithm, beforehand, should determine the optimal sequence length and number of HMMs to employ to achieve some global minima. Given the predictions of HMMs, the algorithm must use them to accurately predict future states. Another challenging task is to update models when necessary, which is aimed to correct the model without incurring large overhead to the monitored infrastructure. We proposed a possible scheme for these issues in this paper.

\subsection{Brief Introduction to Hidden Markov Model}

FHMM uses a Hidden Markov Model (HMM) as its backbone algorithm for learning relationship between defined discrete states. An HMM \cite{rabiner, alghamdi} is a statistical Markov model with hidden states. These hidden states are not directly visible to the observer. The HMM can be completely defined by parameters $A$ (transition matrix), $B$ (observation matrix), $\pi$ (prior probability) as
\begin{equation} \label{eq1}
    \lambda = \theta(A, B, \pi)
\end{equation}
Two assumptions are made by the model. The first, called the Markov assumption, states that the current state is dependent only on the previous state. The second, called the independence assumption, states that the output observation at time $t$ is dependent only on the current state; it is independent of previous observations and states. Given a set of examples from a process, we would be able to estimate the model parameters $\lambda = \theta(A, B, \pi)$ that best describe that process. Then, we could discover the hidden state sequence that was most likely to have produced a given observation sequence. More details can be found in \cite{rabiner}.

\subsection{HMM from Bayesian Perspective}

In the Bayesian approach we assume some prior knowledge about the learning process or structure employed. For the case of HMM, the prior knowledge in encoded in terms of arcs of HMM, and model parameters. This prior knowledge is represented in terms of prior distribution is used to obtain posterior distribution over model structure and parameters. 
More formally, assuming a prior distribution over model structure $P(M)$ and a prior distribution over parameters for each model structure $P(\theta | \text{M})$, a data set $D$ is used to form a posterior distribution over models using Bayes rule (\cite{ghahramani}):
\begin{equation} \label{eq2}
    P(M|D) = \frac{\int P(D | \theta,M)P(\theta | M) d\theta P(M)}{P(D)}
\end{equation}
which averages over uncertainty in parameters. 
The posterior distribution over parameters is computed as:
\begin{equation} \label{eq3}
    P(\theta|M, D) = \frac{P(D | \theta, M)P(\theta | M)}{P(D|M)}
\end{equation}
If we wish to predict the next observation, $Y_{T+1}$ based on our data and models, the Bayesian prediction
\begin{equation} 
    \label{eq4}
    P(Y_{T+1}|D) = \int{P(Y_{T+1}| \theta, M, D)P(\theta | M, D)}{P(M|D)}d\theta dM
\end{equation}
averages over both the uncertainty in the model structure and its parameters. This is known as the predictive distribution for the model \cite{ghahramani}. 

For complex attacker behaviour modelling problems, using sequences with different characteristics to learn one Hidden Markov Model leads to too much generalization and thus losing the discriminating characteristics of the different attackers. Thus, it is evident that the use of  diverse attack sessions, to train a single model, would lead to loss of information which might be unique to a small number of attack sequences. For a threat intelligence model, capturing this information becomes important since it might a crude attack to hack into the system. The proposed approach, called fusion HMM (FHMM), attempts to partition the training data according to the distribution of the number of attack sessions with respect to their lengths, and then train multiple HMMs in a semi-supervised fashion. The predictions of these models are combined using a nonlinear network to provide better, robust predictions. This approach, thus aims to capture the characteristics of the attack sequences that would be lost while using a single global model.

\section{Methodology}
\label{sect:methodology}

Using a single Bayesian structure like HMM to model the joint distribution of all the observations and hidden state, makes optimising $\theta$ to maximise the likelihood intractable. Particularly when the data set $D$ consists of mixture of distributions over different sequence types where the individual distribution might not be a Gaussian in nature, optimising $\theta$ to maximise the likelihood $P(D| \theta, M)$ becomes challenging. Hence we propose segmenting data set D into sub parts such that $D = \big\{D^{(1)}, D^{(2)}, ... D^{(K)}\big\}$ where $D^{(i)}$ are independent and identically distributed observation sets. Empirical methodology for segmenting data set into $K$ parts is provided in this section. The benefits of this segmentation are made apparent in further equations. Intuitively, this is equivalent to breaking down the problem into smaller independent sub-problem of $P(Y_{T+1}|D^{(i)})$, which allows using different $\theta$ values to reach a local minima for that sub-problem. Modifying the original equation to incorporate sub data set:
\begin{equation} 
\label{eq5}
\begin{split}
    \prod_{i=1}^{K} P(Y_{T+1}|D^{(i)}) = \int \prod_{i=1}^{K} & P(Y_{T+1}|\theta , M, D^{(i)})~ P(\theta | M, D^{(i)})~ \\ 
    & P(M|D^{(i)})~d\theta dM
\end{split}
\end{equation}

In the limit of large data set and an uninformative or uniform prior over parameters, the posterior $P(\theta\ |M,D^{(i)})$ will be sharply peaked around the maxima of the likelihood, and therefore the predictions of a single maximum likelihood model will be similar to those obtained by Bayesian integration over parameters.
\begin{equation} \label{eq6}
    \prod_{i=1}^{K} P(\theta|M, D^{(i)}) = \prod_{i=1}^{K} P(D^{(i)}|\theta, M)
\end{equation}

The maximum likelihood (ML) model parameters are obtained by maximising the likelihood or log likelihood.
\begin{equation} \label{eq7}
    \mathcal{L}(\theta, M) = \prod_{i=1}^{K} P(D^{(i)} | \theta, M)
\end{equation}
Further we assume a limiting case of Bayesian approach to learning if we assume a single model structure $M$ and we estimate the parameter vector $\theta$ that maximising the likelihood $P(Y^{(i)} | \theta)$ under that model. If model structure is assumed constant,

\begin{equation} \label{eq8}
    \mathcal{L}(\theta) = \prod_{i=1}^{K} P(D^{(i)} | \theta)
\end{equation}
\begin{equation} \label{eq9}
    \log \mathcal{L}(\theta) = \log \prod_{i=1}^    {K} P(D^{(i)} | \theta)
\end{equation}
\begin{equation} \label{eq10}
    \log \mathcal{L}(\theta) = \log P(D^{(1)} | \theta) + \log P(D^{(2)} | \theta) + .. \log P(D^{(K)} | \theta)
\end{equation}
\begin{equation} \label{eq11}
    \log \mathcal{L}(\theta) = \log \mathcal{L}_{1}(\theta) + \log \mathcal{L}_{2}(\theta) + .. \log \mathcal{L}_{K}(\theta) 
\end{equation}

In order to maximize the overall log likelihood with respect to model $\theta$, we need to maximize the individual log likelihood terms where $\theta$ in each term can be different corresponding to differing model parameters required to aptly represent the joint distribution observation and hidden state.  
\begin{equation} \label{eq12}
    \log \mathcal{L}(\theta) = \sum_{i=1}^{K} \log \mathcal{L}_{i}(\theta_{i})
\end{equation}
These individual models are represented by HMM whose parameters are estimated by Baum Welch algorithm, a special case of EM algorithm. We train $K$ HMM independently on each sub-data using Baum Welch, such that $\lambda^{(i)} = \underset{\theta}{\arg\max}~P( D^{(i)}| \theta)$ where $i \in (0,..K)$. If $Y^{(i)} = \big\{Y^{(i)}_0, Y^{(i)}_1, ... \big\}$ is a sample sequence in $D^{(i)}$ then,

\begin{equation} \label{eq13}
    P(Y^{(i)}_{N+1}|Y^{(i)}_{1:N}, \lambda_i) = \frac{P(Y^{(i)}_{1:N},Y^{(i)}_{N+1}|\lambda_i)}{P(Y^{(i)}_{1:N}|\lambda_i)}
\end{equation}
\begin{equation} \label{eq14}
    P(Y^{(i)}_{N+1}|Y^{(i)}_{1:N}, \lambda) \propto P(Y^{(i)}_{1:N},Y^{(i)}_{N+1}|\lambda_i)
\end{equation}
\begin{equation} \label{eq15}
    X_i = \underset{Y_{N+1}}{\arg\max}~P(Y^{(i)}_{1:N},Y^{(i)}_{N+1}|\lambda_i)
\end{equation}

The denominator being independent from $Y_{N+1}$, we compute $P(Y^{(i)}_{1:N},Y^{(i)}_{N+1}|\lambda_i)$ for each possible $Y_{N+1}$. The value of $Y_{N+1}$ which provides maximum likelihood can be estimated as the best guess for the next observation for model $\lambda_i$. We now have $X_i$ which are estimates obtained from multiple distributions $P(Y_{T+1}|D^{(i)})$ instead of a estimate derived from a single distribution $P(Y_{T+1}|D)$. 

The true estimate given $D = \big\{D^{(1)}, D^{(2)}, ... D^{(K)}\big\}$ is a combination of $X_i$'s which might not be a linear combination. Let function $f$ be used to represent the non linear deterministic component of the approximation, $w_t$ the random noise, then the true output $Y_{T+1}$ is given by:
\begin{equation} \label{eq16}
    Y_{T+1} = f(X_i) + w_t
\end{equation}
Expressing true outcome $Y_{T+1}$ is this form, helps in circumventing time-invariant assumption of state transition and emission matrix of earlier trained HMM $\lambda_{i}$. Therefore the function $f$ along with incorporating the Bayesian analysis of HMM, now incorporates time-step t as input which acts as a deciding factor for providing non-linear weight-age to each $X_i$
\begin{equation} \label{eq17}
    Y_{T+1} = f(X_i,t) + w_t
\end{equation}
We choose to approximate this mapping of $X_i, t$ to $Y_{T+1}$ by neural network.

Thus, the proposed Fusion Hidden Markov Model (FHMM) algorithm can be subdivided into three major steps -

\begin{enumerate}
    \item Finding optimum value of $K$ according to the distribution of input sequences and clustering data into $K$ diverse groups.
    \item Training $K$ different Hidden Markov Models on the $K$ subgroups.
    \item Learning mapping of individual HMM predictions to a single prediction using a Neural Network.
\end{enumerate}

\noindent The basic terminology used in this paper is as follows - \\
\hspace*{4mm}$T =$ length of the observation sequence \\ 
\hspace*{4mm}$N =$ number of states in the model \\ 
\hspace*{4mm}$M =$ number of observation symbols \\ 
\hspace*{4mm}$Q =$ distinct states of Markov process $= \big\{q_0, q_1, .., q_{N-1}\big\}$ \\ 
\hspace*{4mm}$V =$  set of possible observations $= \big\{0, 1, . . . , M - 1\big\}$ \\ 
\hspace*{4mm}$\pi =$ initial state distribution \\
\hspace*{4mm}$A =$ state transition probabilities 
$= {a_{ij}}$ with shape $N$x$N$ \\
\hspace*{9mm} where $a_{ij} =$ $P(\text{state}~q_j~\text{at}~t+1~|~\text{state}~q_i~\text{at}~t)$ \\ 
\hspace*{23.5mm}$=~P(x_{t+1}|x_t)$\\ 
\hspace*{4mm}$B =$ observation probability matrix \\ 
\hspace*{7mm}$= \{b_j(k)\}$ with shape $N$x$M$ \\
\hspace*{10mm}where $b_j(k) = P(\text{observation}~k~\text{at}~t~|~\text{state}~q_j~\text{at}~t)$ \\ 
\hspace*{26.5mm}$= P(y_t|x_t)$ \\ 
\hspace*{4mm}$O = (O_0, O_1, . . . , O_{T−1})$ \\ 
\hspace*{7.5mm}$=$ observation sequence \\
\hspace*{10mm}where $O_i~\in~V_x~$ $\text{for}~i \in \{0, 1, ..., T-1\}$

\noindent An HMM is completely defined by parameters $A$, $B$, $\pi$ and denoted by $\lambda = \theta(A, B, \pi)$.

\subsection{Partitioning Data into $K$ Groups}

An important part of any ensemble learning algorithm is to partition data into non-correlated subsets which will make learning from it fruitful. The non-correlated $K$ subsets of data also make sure HMM’s are fitted to different information subset which when combined will characterise the whose time-series data. The number $K$ is a hyperparameter which characterizes the complexity of the data. There exists a tradeoff between $K$ used and time required to train HMM as complexity of data increases. To satisfy the above conditions FHMM uses a dissimilarity function $f(F)$ to divide data into $K$ subsets such that each subset captures a particular pattern of temporal data. The dissimilarity function is a custom defined distance function whose definition would change depending on the type of data used. Initially, dividing data depending upon length of sequence provides two benefit- 

\begin{itemize}
    \item The data of similar length generally originates from same distribution. To illustrate, in case of cyber attacks, the attacks of similar length generally employ same attack strategy and pattern. This makes sure that we are capturing attacker mindset or bot attack patterns in subset of data. This will help HMM to learn the pattern quickly with higher accuracy. 
    \item From implementation and operational point of view. HMMs train on data of same length comparatively faster as compared to data consisting of variable length.
    \item Diverse state transitions are present in attack sequences of significantly different lengths. Thus, it becomes difficult for a single HMM to accurately model it.
\end{itemize}

\begin{algorithm}
\caption{Choosing and Training HMM}
\label{alg1}
\begin{algorithmic}[1]
    \renewcommand{\algorithmicrequire}{    \hspace*{1.5\algorithmicindent}\vspace*{0.1\baselineskip}\textbf{Input:}}
    \renewcommand{\algorithmicensure}{    \hspace*{1.5\algorithmicindent}\vspace*{0.1\baselineskip}\textbf{Output:}}
    \REQUIRE $j=[O_0, O_1, …. O_i], N, M$, \\ 
    \hspace*{10mm}Dissimilarity function $(f(F))$
    \ENSURE $[\lambda_0, \lambda_1, …. \lambda_B]$
    \STATE $L$ = max length ($j$)
    \FOR {all 0 in $j$}
        \STATE $D_i = O$ where $i \in \text{int}(0,L)$
    \ENDFOR
    \FOR {all $D_i$}
        \STATE Compute Frequency array $F[i]$ where $i \in V$
    \ENDFOR
    \STATE Construct similarity sets $S$ by computing Euclidean distance between frequency arrays $F[i]$
    \STATE Assign Ranks $R_i$, where $i \in V$ using $S$
    \STATE Use $R_i$ to obtain lengths $l_i$ 
    \RETURN $l_i$
\end{algorithmic}
\end{algorithm}

The data partitioned into different subsets by length is then chosen for the ensemble depending on its correlation with each other. For each sub dataset $D_i$ where $i~\epsilon~int(0, max(l))$ we compute a frequency occurrence array. This frequency array is computed for each sub dataset $D_i$. The frequency array consists of occurence of discrete states in the sub dataset $D_i$. This frequency occurrence array consists of occurence of each state divided by total states seen. The frequency array characterises the sub-dataset $D_i$ quantitatively which makes it possible to compare its ‘similarity’ with other $D_i$. The $G$ sub dataset $D_i$ is represented by a $1 \times N$ frequency array vector where $N$ is number of discrete states. Then the $G \times N$ frequency matrix is used to obtain a set of  datasets $D_j$ called similarity set, which is  similar to a particular dataset $D_i$ by computing euclidean distance between the corresponding frequency arrays. These similar sets are used to rank datasets $D_i$ to select $K$ datasets $D_i$ from total $G$ sub dataset $D_i$ such that these $K$ datasets are diverse, have low correlation with each other and cover most of the training data.

\begin{algorithm}
\caption{Individual Hidden Markov Model Training and Updating Process}
\label{alg2}
\begin{algorithmic}[1]
    \renewcommand{\algorithmicrequire}{    \hspace*{1.5\algorithmicindent}\vspace*{0.1\baselineskip}\textbf{Input:}}
    \renewcommand{\algorithmicensure}{    \hspace*{1.5\algorithmicindent}\vspace*{0.1\baselineskip}\textbf{Output:}}
    \REQUIRE $O, N, M$
    \ENSURE HMM Parameters $\lambda = (A, B, \pi)$
    \STATE Random Initialize $\lambda = (A, B, \pi)$
    \STATE Forward pass: Compute forward probabilities $\alpha_{t}(i)$  \\ 
    Normalize computed $\alpha_{t}(i)$
    \STATE Backward pass: Compute backward probabilities $\beta_t(i)$ \\ 
    Normalize computed $\beta_t(i)$
    \STATE Compute Di-gammas and Gammas
    \STATE Re-estimate $A, B,$ and $\pi$
    \STATE Compute $\log [P(O|\lambda)]$
    \IF {$\log[P(O|\lambda)]_{\text{new}} > \log[P(O|\lambda)]_{\text{old}}$} 
        \STATE Go to step 2
    \ELSE 
        \RETURN $\lambda = (A, B, \pi)$
    \ENDIF
    \STATE End
\end{algorithmic}
\end{algorithm}

\subsection{Training K HMMs}

In the second step, a group of attack sequences in the training data $D_i$, is used to learn an HMM model $\lambda_i$. Although the use of small set of attack sequences to learn an HMM $\lambda_i$ might lead to overfitting, in a broader context, it ensures that each model captures the set of characteristics distinct to that group of sequences. Each Hidden Markov Model $\lambda_i$ is trained using Baum-Welch algorithm \cite{rabiner} along with Expectation Maximization step (EM) \cite{moon} in order to train the parameters (transition, emission and prior probabilities) of the model. EM maximizes the likelihood of each sequence with respect to the corresponding model $\lambda_i$ i.e., $P(D_i|\lambda_i)$. Thus, each model is capable of accurately predicting future steps in a sequence which belongs to the probability distribution learnt by it. The detailed training procedure is given in Algorithm~\ref{alg2}. In each model the number of hidden states can be viewed a hyper-parameter which can be tuned.

\subsection{Combining Predictions of $K$ HMMs}

The resulting predictions from $K$ Hidden Markov Models are then aggregated by neural network to generate a combined prediction for next step. The neural network layer consists of a linear weight component followed by a nonlinear function. It minimizes the mean squared error of the HMM predictions on the training data and assigns a probabilistic weight to each HMM. Thus, the neural network layer learns a nonlinear mapping from the predictions of HMMs to the next state output. 
The $K$ Hidden Markov Model predictions are then assigned a weight $W$ where equation (summation of $W$ over all $K$ HMM $= 1$). This importance to particular HMM’s output characterized by weight $W$ is learnt by iterating through samples.

\begin{algorithm}
\caption{Individual Hidden Markov Model Prediction Process}
\label{alg3}
\begin{algorithmic}[1]
    \renewcommand{\algorithmicrequire}{    \hspace*{1.5\algorithmicindent}\vspace*{0.1\baselineskip}\textbf{Input:}}
    \renewcommand{\algorithmicensure}{    \hspace*{1.5\algorithmicindent}\vspace*{0.1\baselineskip}\textbf{Output:}}
    \REQUIRE $O, \lambda, V$
    \ENSURE $X_{k}(t)$
    \STATE Initialize: $U$ array
    \FOR {$k$ in $V$}
        \STATE $O$.append($V, k$)
        \STATE Compute Gamma
        \STATE $U$.append($U_k$)
    \ENDFOR
    \RETURN argmax($U$)
    \STATE Train individual HMM on $D_i$ where $i \epsilon l_i$
    \RETURN [$\lambda_0, \lambda_1, …. \lambda_B$]
\end{algorithmic}
\end{algorithm}

\begin{figure}[H]
\centering
    \includegraphics[width=0.3\textwidth,keepaspectratio]{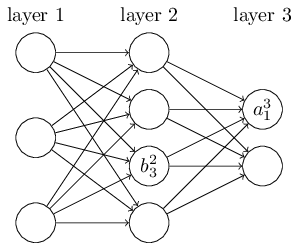}
    \caption{Mapping estimates using neural network}
    \label{fig:fig1}
\end{figure}

The problem is now defined as given predictions [$X_0, X_1, …. X_K$] from HMM $[\lambda_0, \lambda_1, …. \lambda_K]$ at timestep $t = T-1$, what is the likely action the attacker will take at timestep $t=T$.
FHMM uses supervised neural network, as shown in Figure~\ref{fig:fig1}, for learning the weight $W$ in training phase from $N$ data samples and then uses the learned weight $W$ to estimating the future action taken. The network used is characterized by -
\begin{equation} \label{eq18}
    \alpha_j^l = \sigma(\sum_{k}(w_{jk}^l\alpha_k^{l-1}+b_j^l)
\end{equation}
where $a_j^l$ is the activation and $b_j^l$ of the jth neuron in the lth layer.

We define a cost $C$ which is used as a measure to indicate the offset of predicted action from actual action taken. We define the cost $C$ over all samples $n$ of data as:
\begin{equation} \label{eq19}
    C = \frac{1}{2n}\sum_{x}||y(x) - \alpha^L(x)|| 
\end{equation}
This equation characterises the difference in the true output $y(x)$ and the output predicted by network $y'= a^L(x)$. Then we use backpropagation (\cite{hinton}) to compute the gradients and update the weights of the network.

The depth of network required in terms of layers $L$ increases as the complexity of pattern increases. Apart from neural networks superior ability to model nonlinear complex relationship in data, it has following benefits over traditional parametric approaches - 

\begin{itemize}
    \item Neural Network generalize well to unseen data and can infer unseen patterns not initially present in data provided in training. This is extremely crucial in cyber security applications where new attacks consists of previously employed strategies intermittently spread throughout the attack. The neural network can detect this type of sub-pattern in attack and therefore complements the ability of HMM.
    \item Unlike HMM and other parametric techniques, neural network does not impose restriction on the distribution of the input variables. Moreover, neural networks can better model heteroscedasticity, while traditional model fail in to model data with high volatility and non-constant variance which is common in cyber security applications.
\end{itemize}

\noindent The training and evaluating phase is given in Algorithm~\ref{alg4}.

\begin{algorithm}
\caption{Second Stage Data Collection}
\label{alg4}
\begin{algorithmic}[1]
    \renewcommand{\algorithmicrequire}{    \hspace*{1.5\algorithmicindent}\vspace*{0.1\baselineskip}\textbf{Input:}}
    \renewcommand{\algorithmicensure}{    \hspace*{1.5\algorithmicindent}\vspace*{0.1\baselineskip}\textbf{Output:}}
    \REQUIRE $O,$ [$\lambda_0, \lambda_1, ..., \lambda_B$]
    \ENSURE $X_{k}$
    \FOR {$k$ in $n$}
        \STATE $X_k$ = Predictions($\lambda, O, V$)
    \ENDFOR
\end{algorithmic}
\vspace{1mm}
\textbf{Training phase}:
\begin{algorithmic}[1]
    \renewcommand{\algorithmicrequire}{    \hspace*{1.5\algorithmicindent}\vspace*{0.1\baselineskip}\textbf{Input:}}
    \renewcommand{\algorithmicensure}{    \hspace*{1.5\algorithmicindent}\vspace*{0.1\baselineskip}\textbf{Output:}}
    \REQUIRE [$X_0, X_1, ..., X_k$], $Y, lr = \text{learning rate}$
    \ENSURE $f (x; \theta, w)$
    \STATE Initialize: $W, w$ randomly
    \STATE $Y' = w^T.\max(0, W^{T}x+c) + b$
    \STATE Compute $C =$ Cost($Y, Y'$)
    \STATE Update parameters:
        \\ $W_{\text{new}} = W_{\text{old}} - \Delta C.lr$ \\
    \IF {Cost($Y_{\text{new}}, O$) $<$ Cost($Y_{\text{old}}, O$)}
        \STATE Go to step 1
    \ELSE
        \RETURN $f (x; \theta ,w)$
    \ENDIF
\end{algorithmic}
\vspace{1mm}
\textbf{Evaluation phase}:
\begin{algorithmic}[1]
    \renewcommand{\algorithmicrequire}{    \hspace*{1.5\algorithmicindent}\vspace*{0.1\baselineskip}\textbf{Input:}}
    \renewcommand{\algorithmicensure}{    \hspace*{1.5\algorithmicindent}\vspace*{0.1\baselineskip}\textbf{Output:}}
    \REQUIRE [$X_0, X_1, ..., X_k$]
    \ENSURE $Y'$
    \STATE $Y' = w^{T}.\max(0, W^{T}x+c) + b$
\end{algorithmic}
\end{algorithm}

\subsection{Overall Algorithm}

During training stage, FHMM algorithm takes multiple attack sequence consisting of discrete states as input. Depending on sequence length and rank computed using dissimilarity function it selects and divides data into least correlated $K$ sub data. Then the predictions obtained from $K$ HMM trained to overfit on the $K$ sub data are fed to an neural network for learning the weightage of each HMM. 

During prediction phase, the intermediate HMM predictions are fed to neural network which outputs a single value of next state in sequence of attack. Figure~ref{fig:fig2} illustrates the training and prediction phases in FHMM algorithm.
\begin{figure}[H]
\centering
    \includegraphics[width=0.45\textwidth]{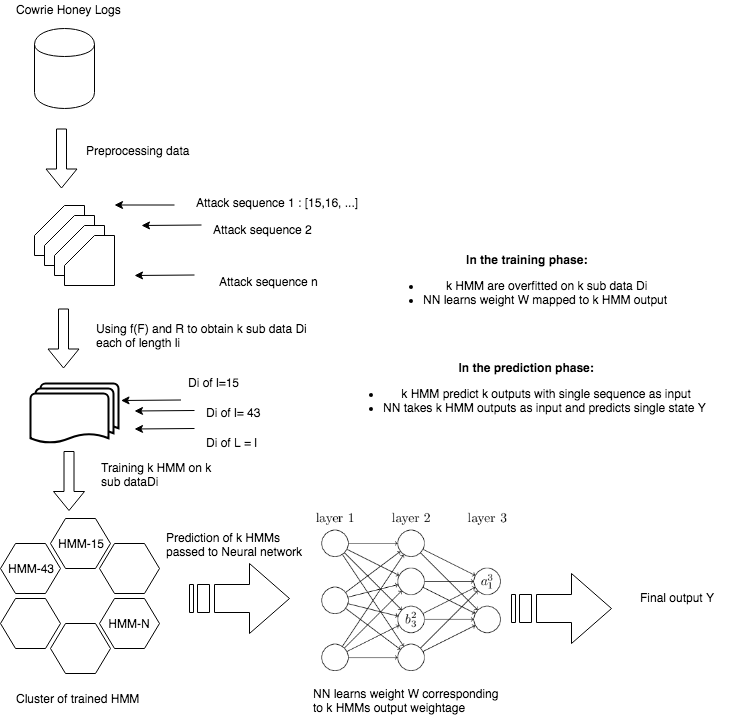}
    \caption{Overall FHMM Algorithm}
    \label{fig:fig2}
\end{figure}

\section{Experiments and Discussion}
\label{sect:exp}

\subsection{Dataset Used}
For collecting real attack logs, we had setup Cowrie honeypot \cite{cowrie} which is a medium interaction SSH and telnet honeypot. Honeypot is a decoy system with the sole intention of tricking attacker with an easy target to log his attack patterns. In case of Cowrie, the attack patterns are logged in JSON format. The detailed description of  attack features logged and dataset description is provided by \cite{rade}. We processed events in Cowrie logs and divided them into 19 commands consisting of: 
\begin{center}
\textit{client.size, client.version, command.failed, command.input/delete, command.input/dir-sudo, command.input/other, command.input/system, command.input/write, command.success, direct-tcpip.data, direct-tcpip.request, log.closed, log.open, login.failed, login.success, session.closed, session.connect, session.file-download, session.input}
\end{center}

\begin{figure*}[t]
    \begin{minipage}[c]{.32\textwidth}
        \centering
        \includegraphics[width=\linewidth,height=97pt]{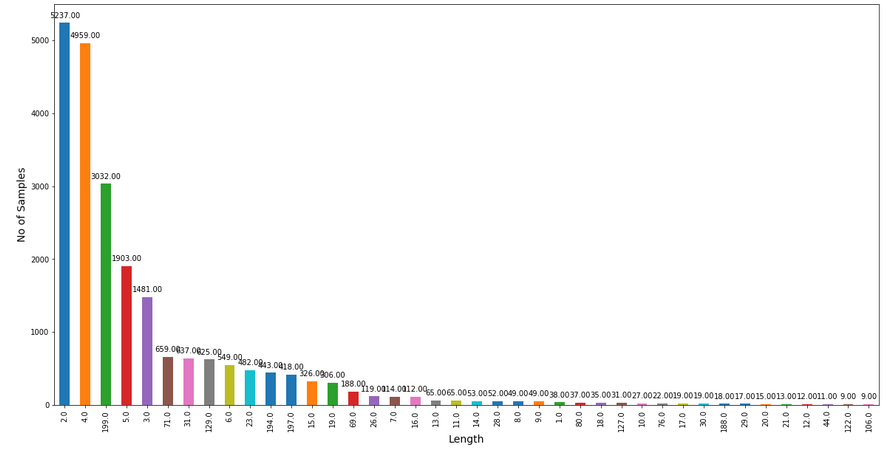}
        \caption{Distribution of attacker sessions.}
        \label{fig:fig3}
    \end{minipage}
    \hspace{1mm}
    \begin{minipage}[c]{.32\textwidth}
        \centering
        \includegraphics[width=\linewidth,height=97pt]{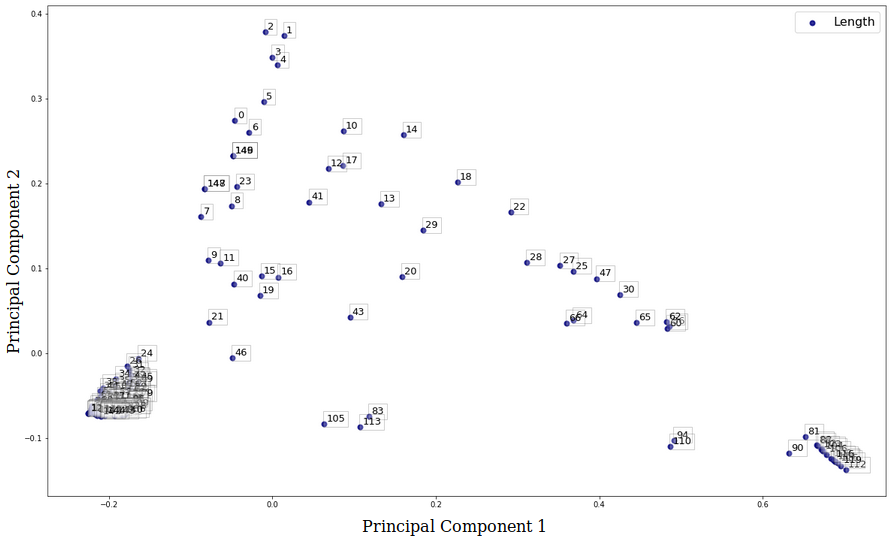}
        \caption{2-D plot of frequency arrays.}
        \label{fig:fig4}
    \end{minipage}
    \hspace{1mm}
    \begin{minipage}[c]{.32\textwidth}
        \vspace*{3mm}
        \centering
        \includegraphics[width=\linewidth,height=97pt]{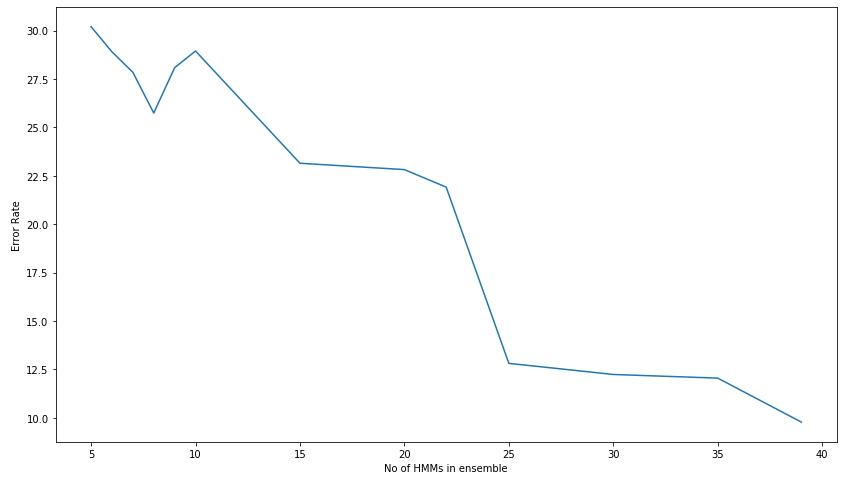}
        \caption{FHMM error rate vs number of HMMs.}
        \label{fig:fig5}
    \end{minipage}
\end{figure*}


We encoded these events as discrete states labelled $0$ to $18$. These $19$ states were used for modelling and prediction in HMM, LSTM and FHMM algorithm. The data is grouped by session id for considering each sequence where each session id corresponds to the sequence of actions taken by hacker. The assumption made here is different session id are independent of individual attacker characteristics and hence dividing depending on session ID rather than source IP would not affect the modelling and prediction by a large factor.

The dataset used for training FHMM was extracted from the logs generated by Cowrie honeypot from April 2017 to July 2017. By processing these logs, we generated 22,499 distinct attack sessions involving the sequence of steps taken by a particular source IP. The attack sessions lasted from 2 steps to over 1400 steps. These sessions are raw logs of shell interaction performed by the attacker. We evaluate the performance of FHMM on a separate test set comprising of real time logs of attackers’ actions for 1 month.

\subsection{Partitioning Data into $K$ groups}

In cyber security logs, there are multiple attacks from different hackers of variable attack lengths. This results in logs being generated which are varied in terms of types, pattern, sequence length, and duration of attack or infiltration. The data can be divided into sub-datasets depending on either of the factors, each with its merits and demerits. Due to phenomena prevalent in cyber security domain where attacks of similar length following similar patterns, we have considered the distribution of the number of attack sessions with respect to their lengths and partitioned the preprocessed training data according to lengths. Figure~\ref{fig:fig3} shows the distribution of number of attack sessions with respect to their lengths.



Once the data is divided into sub-datasets we can train $N$ number of HMM on the sub-datasets and pass their predictions to Neural Network. But to provide fast response time to prevent real-time attack, training and predictions of $N$ HMMs where $N$ can be in thousands is not practically feasible. Hence we use similarity measures to reduce the sub-datasets from $N$ to $K$. In order to divide training data into $K$ sets, we construct 19-dimensional frequency arrays $F[i]$ for each dataset $D_i$ consisting of sequences of length $i$ where $1 \leq i \leq L$. The frequency arrays describe the distribution of different states in terms of probability of occurrence of each state across multiple attacks of  same length. Figure~\ref{fig:fig4} shows these frequency arrays reduced to  2-dimensional vector space by employing PCA \cite{shlens}. Using frequency array $F[i]$ to characterise all attacks of particular length, is sufficiently accurate to compute dissimilarity  measure between attacks of different lengths. It is also essential the sub-dataset $K$ produced from $N$ be containing maximum information along with being mutually exclusive and dissimilar with one another. Euclidean distance between these arrays along each of 19 dimension provides the dissimilarity between the arrays. Hence we compute euclidean distance between these arrays $F[i]$ to find dissimilar sets and select $K$ datasets such that these $K$ datasets cover maximum information present in the training data. The selected $K$ datasets out of $N$ total datasets helps in reducing training and prediction time without significantly affecting accuracy of deployed system.

\begin{figure*}[t]
    \begin{minipage}[c]{.32\textwidth}
        \centering
        \includegraphics[width=\linewidth,height=97pt]{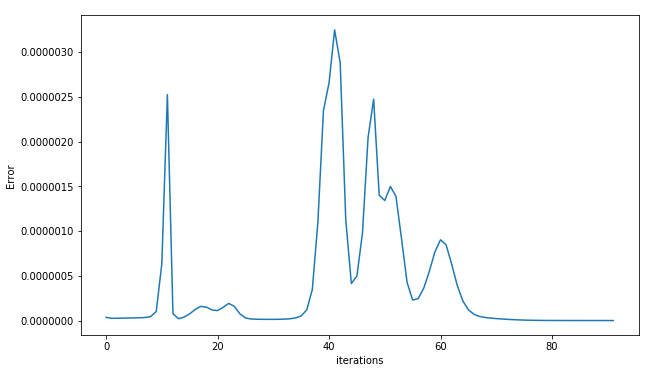}
        \caption{Error rate while training a single HMM.}
        \label{fig:fig6}
    \end{minipage}
    \hspace{1mm}
    \begin{minipage}[c]{.32\textwidth}
        \centering
        \includegraphics[width=\linewidth,height=97pt]{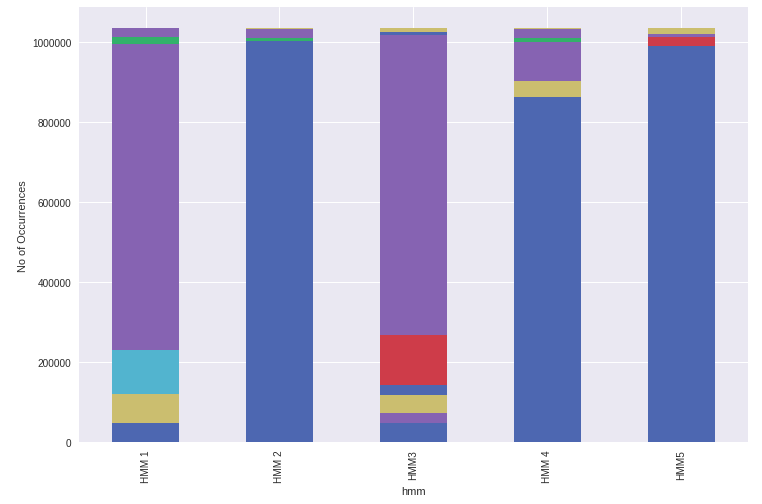}
        \caption{State distribution of different HMMs.}
        \label{fig:fig7}
    \end{minipage}
    \hspace{1mm}
    \begin{minipage}[c]{.32\textwidth}
        \centering
        \includegraphics[width=0.8\linewidth,height=97pt]{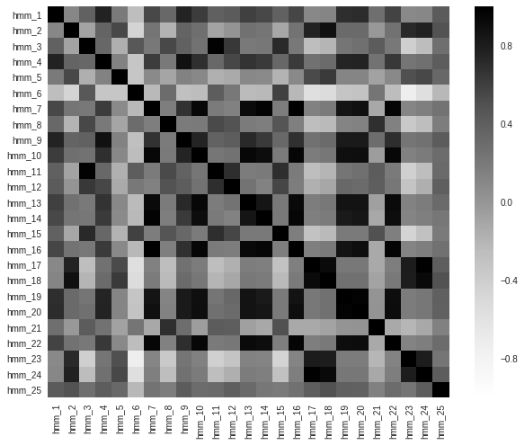}
        \caption{Correlation between HMM's predictions.}
        \label{fig:fig8}
    \end{minipage}
\end{figure*}



The Figure~\ref{fig:fig6} depicts individual HMM error curve while training.

\subsection{Training $K$ HMMs}
The $K$ datasets selected, are used to train $K$ different HMMs This HMMs are trained until convergence using Baum Welsh algorithm. HMM is implemented in cython and trained parallely, thus reducing computation time by a large factor. Then each of the $K$ HMMs predicts the next possible state, resulting in $K$ predictions for next state which are fed to neural network.

To illustrate the practical need of splitting data and training $K$ different HMM, see the Figures~\ref{fig:fig4},~\ref{fig:fig7}~and~\ref{fig:fig8}. The Figure~\ref{fig:fig7} shows HMM trained on different lengths and the distribution of output states predicted by HMM. Here each state type is represented by different color and clearly shows why different $K$ HMMs are required to model entire data. Moreover, it also depicts that each HMM is learning different data patterns unique to an attacking type. 



\subsection{Combining Predictions of $K$ HMMs}
The predictions of these HMMs are combined by a neural network containing 60 units with ReLU activation. Selecting the optimum value of $K$ is a classical bias-variance tradeoff. Large values of $k$ ($>$50) lead to poor generalization on test data. Table~\ref{tab:tab1} shows the weight assigned to each feature input given to neural network in descending order. Here the features are prediction made by HMM of trained on particular length. The table clearly indicates temporal feature count has the highest importance in determining the next state taken by attacker, followed by predictions from HMM trained on length 11 and 44.

\subsection{Quantitative Results}

Selecting the optimum value of $K$ is a tradeoff between error rate and computational requirements. Increasing the value of $K$ reduces the error rate. Figure~\ref{fig:fig5} shows a plot of error rate vs number of models in FHMM.

As evident, there is a  significant reduction in error due to adding learners to the fusion. For our dataset, much of the reduction appears after 20-25 classifiers. One reason for this is the diverse set of features learnt by HMMs - 23, 24 and 25. This is evident from the correlation between the predictions of individual HMMs as shown in Figure~\ref{fig:fig8}.

\begin{table}[H]
\caption{Weights Assigned by Neural Network to Individual HMMs}
\label{tab:tab1}
\footnotesize
\centering
\begin{tabular}{c|c}
    \hline
    \bfseries Weight & \bfseries Feature \\
    \hline
    12.0151 $\pm$ 0.0308 & count  \\
    4.6469 $\pm$ 0.0179 & hmm\_11.0  \\
    3.9593 $\pm$ 0.0131 & hmm\_44.0  \\
    3.9512 $\pm$ 0.0261 & hmm\_13.0  \\
    2.9618 $\pm$ 0.0069 & hmm\_188.0  \\
    2.6922 $\pm$ 0.0071 & hmm\_106.0  \\
    2.6831 $\pm$ 0.0089 & hmm\_23.0  \\
    2.6451 $\pm$ 0.0069 & hmm\_127.0  \\
    2.5365 $\pm$ 0.0214 & hmm\_28.0  \\
    2.3346 $\pm$ 0.0166 & hmm\_18.0  \\
    2.2671 $\pm$ 0.0220 & hmm\_76.0  \\
    2.1762 $\pm$ 0.0204 & hmm\_129.0  \\
    2.0389 $\pm$ 0.0093 & hmm\_199.0  \\
    1.7186 $\pm$ 0.0198 & hmm\_4.0  \\
    1.6456 $\pm$ 0.0043 & hmm\_80.0  \\
    1.5216 $\pm$ 0.0082 & hmm\_9.0  \\
    1.4252 $\pm$ 0.0136 & hmm\_6.0  \\
    1.3633 $\pm$ 0.0047 & hmm\_20.0  \\
    1.3594 $\pm$ 0.0037 & hmm\_69.0  \\
    1.2440 $\pm$ 0.0063 & hmm\_194.0  \\
    1.2393 $\pm$ 0.0072 & hmm\_197.0  \\
    1.1861 $\pm$ 0.0085 & hmm\_14.0  \\
    1.1355 $\pm$ 0.0061 & hmm\_31.0  \\
    \hline
\end{tabular}
\end{table}

After 35-40 models, the error reduction for FHMM appears to have nearly reached a plateau.  So, we primarily focus on the performance of FHMM for $K$ = 38. Table~\ref{tab:tab2} shows the prediction accuracy attained by FHMM with $K$ = 25 and $K$ = 38 along with that of  other sequence models such as Markov chain, single HMM and LSTM. Depending on how the HMMs in FHMM are trained, the training time differs. As the training of individual HMM is independent of other HMMs, this allows parallel training with faster training times compared to sequential training pipeline.


\begin{table}[b]
\caption{Comparison of accuracy obtained by different models}
\label{tab:tab2}
\footnotesize
\centering
\begin{tabular}{c|c|c}
    \hline
    \bfseries Model & \bfseries Accuracy & \bfseries  Training Time (in hrs)\\
    \hline
    Markov Chain & 72 & 0.3  \\
    HMM & 77 & 1\\
    LSTM & 86 & 5 \\
    FHMM (K=25, sequential) & 87.19 & 2.3 \\
    FHMM (K=38, sequential) & 90.82 & 2.5 \\
    FHMM (K=25, parallel) & 87.19 & 1.3 \\
    FHMM (K=38, parallel) & 90.82 & 1.5 \\
    \hline
\end{tabular}
\end{table}

One obvious conclusion drawn from the results is that the reduction in error rate provided by FHMM is very large as compared to that of a single learner. Additionally, FHMM has a better generalization ability than single models which may be attributed to the following reasons. The training data contains considerably diverse information and it becomes difficult for a single learner to learn a generalized joint probability distribution over the data. Thus, we use many learners which perform well on parts of data. These learners may learn different distributions over the data and combining them is a convenient choice. Experimentally Table~\ref{tab:tab3} illustrates this result, where accuracy of each state as predicted by HMM trained on different lengths is compared with FHMM. Training many learners also circumvents the imperfect search process of HMM. In HMM, we assume some prior knowledge about the learning process and the model structure and  the desired complex input-output mapping may not be present in the hypothesis space being searched by the learning algorithm. In such cases, exploiting multiple learners provides a better estimate.

\begin{table*}[!ht]
\caption{Accuracy obtained for different states by FHMM (k=38)}
\label{tab:tab3}
\tiny
\centering
\begin{tabular}{c | c | c | c | c | c | c | c | c | c | c | c | c | c | c | c | c | c | c | c}
    \hline \bfseries State &
    \bfseries 0 & \bfseries 1 & \bfseries 2 & \bfseries 3 & \bfseries 4 & \bfseries 5 & \bfseries 6 & \bfseries 7 & \bfseries 8 & \bfseries 9 & \bfseries 10 & \bfseries 11 & \bfseries 12 & \bfseries 13 & \bfseries 14 & \bfseries 15 & \bfseries 16 & \bfseries 17 & \bfseries 18 \\
    \hline FHMM &
    0 & 0 & 59.1 & 70.2 & 92.9 & 68.2 & 0 & 97.2 & 99.6 & 59 & 54.3 & 53.6 & 99.1 & 41.6 & 37.3 & 42.4 & - & 25.6 & 42.9 \\
    \hline HMM 9 &
    0 & 100 & 0 & 0 & 0 & 0 & 0 & 0 & 0 & 63.2 & 35.5 & 0 & 0 & 0 & 37.2 & 0 & - & 0 & 0 \\
    \hline HMM 11 &
    0 & 0 & 82.9 & 66.3 & 0 & 0 & 0 & 0 & 2.1 & 0 & 0 & 0 & 97.5 & 0 & 62.5 & 0 & - & 0 & 0 \\
    \hline HMM 44 & 
    0 & 100 & 0.1 & 12.7 & 0 & 0 & 0 & 0 & 6.3 & 0 & 9.7 & 13.2 & 0.8 & 0 & 37.2 & 0.1 & - & 7.1 & 0 \\
    \hline HMM 71 & 
    0 & 0 & 53.4 & 35.3 & 0 & 72.4 & 0 & 17.3 & 81.3 & 0 & 0 & 9.4 & 0 & 0 & 0 & 40.8 & - & 0 & 0 \\
    \hline HMM 199 & 
    0 & 0 & 0 & 73 & 73.2 & 0 & 0 & 0 & 96.9 & 0 & 0 & 0 & 97.5 & 0 & 62.5 & 0 & - & 0 & 0 \\
    \hline
\end{tabular}
\end{table*}

\subsection{Limitations}
Although FHMM algorithm is robust to noise and provides a significant reduction in error rate while modelling attack sequences, it has some limitations common to other ensemble methods. The basic requirement of FHMM is that the base HMMs should be diverse and must have low correlation with each other for a significant reduction in error rate over train distribution. However, creating diverse base models is not always possible. Moreover, it requires the use of techniques such as partitioning data into diverse groups and initializing the base learners differently to induce heterogeneity. In addition, FHMM is complex and computationally expensive as compared to simple probabilistic algorithms such as Markov chain and HMM. With FHMM, learning time and memory constraints need to be taken care of. 

\subsection{Other Applications}
The proposed FHMM algorithm can be easily extended to other sequence problems where the goal is to predict the next state in the sequence. While the hidden state and the observation spaces are discrete in the above FHMM algorithm, the FHMM can also be used to model continuous observations. Potential applications include stock prices prediction, speech synthesis, time-series analysis, gene prediction and parts-of-speech tagging. For these applications, the major portion of the algorithm would remain identical with a change in the criteria for partitioning data into $K$ groups. For this purpose, other methods such as clustering and similarity measures like cosine distance can be employed depending on the training data and the application. After incorporating a suitable partitioning technique, the FHMM algorithm can be identically applied to other sequencing tasks.

\section{Conclusion}
\label{sect:conclusion}
This paper proposes Fusion Hidden Markov Model which exploits the benefit of ensemble learning for modelling behavioural aspect of attacker to obtain better insight on predicting his future actions. FHMM provides compelling results while modelling temporal patterns due to its higher modelling capacity, robustness to noise, and reduced training time. FHMM's superiority is substantiated by comparing against traditional approaches of Markov Chain, HMM, deep LSTM. The model is evaluated on Cowrie Honeypot dataset which consists of large number of diverse real-time attack sessions. Keeping initial conditions and preprocessing constant, the proposed architecture outperforms other traditional and benchmark models. In addition, we explored FHMM in depth, with highlights to individual parameter contribution to the overall model. The architecture of FHMM allows it to be generalized to other domains which involves temporal modelling of sequential data. 

\bibliographystyle{IEEEtran}

\bibliography{main}

\begin{thebibliography}{10}
\providecommand{\url}[1]{#1}
\csname url@samestyle\endcsname
\providecommand{\newblock}{\relax}
\providecommand{\bibinfo}[2]{#2}
\providecommand{\BIBentrySTDinterwordspacing}{\spaceskip=0pt\relax}
\providecommand{\BIBentryALTinterwordstretchfactor}{4}
\providecommand{\BIBentryALTinterwordspacing}{\spaceskip=\fontdimen2\font plus
\BIBentryALTinterwordstretchfactor\fontdimen3\font minus
  \fontdimen4\font\relax}
\providecommand{\BIBforeignlanguage}[2]{{%
\expandafter\ifx\csname l@#1\endcsname\relax
\typeout{** WARNING: IEEEtran.bst: No hyphenation pattern has been}%
\typeout{** loaded for the language `#1'. Using the pattern for}%
\typeout{** the default language instead.}%
\else
\language=\csname l@#1\endcsname
\fi
#2}}
\providecommand{\BIBdecl}{\relax}
\BIBdecl

\bibitem{razzaq}
A.~{Razzaq}, A.~{Hur}, H.~F. {Ahmad}, and M.~{Masood}, ``Cyber security:
  Threats, reasons, challenges, methodologies and state of the art solutions
  for industrial applications,'' in \emph{2013 IEEE Eleventh International
  Symposium on Autonomous Decentralized Systems (ISADS)}, March 2013, pp. 1--6.

\bibitem{gemalto}
\BIBentryALTinterwordspacing
Gemalto, ``Data breach statistics by year, industry, more.'' [Online].
  Available: \url{http://www.breachlevelindex.com/}
\BIBentrySTDinterwordspacing

\bibitem{modi}
\BIBentryALTinterwordspacing
C.~N. Modi and K.~Acha, ``Virtualization layer security challenges and
  intrusion detection/prevention systems in cloud computing: a comprehensive
  review,'' \emph{The Journal of Supercomputing}, vol.~73, no.~3, pp.
  1192--1234, Mar 2017. [Online]. Available:
  \url{https://doi.org/10.1007/s11227-016-1805-9}
\BIBentrySTDinterwordspacing

\bibitem{mukherjee}
B.~{Mukherjee}, L.~T. {Heberlein}, and K.~N. {Levitt}, ``Network intrusion
  detection,'' \emph{IEEE Network}, vol.~8, no.~3, pp. 26--41, May 1994.

\bibitem{viegas}
E.~{Viegas}, A.~O. {Santin}, A.~{França}, R.~{Jasinski}, V.~A. {Pedroni}, and
  L.~S. {Oliveira}, ``Towards an energy-efficient anomaly-based intrusion
  detection engine for embedded systems,'' \emph{IEEE Transactions on
  Computers}, vol.~66, no.~1, pp. 163--177, Jan 2017.

\bibitem{dietterich}
T.~G. Dietterich, ``Ensemble methods in machine learning,'' in \emph{Multiple
  Classifier Systems}.\hskip 1em plus 0.5em minus 0.4em\relax Berlin,
  Heidelberg: Springer Berlin Heidelberg, 2000, pp. 1--15.

\bibitem{cardenas}
A.~A. {Cárdenas}, P.~K. {Manadhata}, and S.~P. {Rajan}, ``Big data analytics
  for security,'' \emph{IEEE Security Privacy}, vol.~11, no.~6, pp. 74--76, Nov
  2013.

\bibitem{alqurishi}
M.~{Al-Qurishi}, M.~S. {Hossain}, M.~{Alrubaian}, S.~M.~M. {Rahman}, and
  A.~{Alamri}, ``Leveraging analysis of user behavior to identify malicious
  activities in large-scale social networks,'' \emph{IEEE Transactions on
  Industrial Informatics}, vol.~14, no.~2, pp. 799--813, Feb 2018.

\bibitem{vallis}
\BIBentryALTinterwordspacing
O.~Vallis, J.~Hochenbaum, and A.~Kejariwal, ``A novel technique for long-term
  anomaly detection in the cloud,'' in \emph{Proceedings of the 6th USENIX
  Conference on Hot Topics in Cloud Computing}, ser. HotCloud'14.\hskip 1em
  plus 0.5em minus 0.4em\relax Berkeley, CA, USA: USENIX Association, 2014, pp.
  15--15. [Online]. Available:
  \url{http://dl.acm.org/citation.cfm?id=2696535.2696550}
\BIBentrySTDinterwordspacing

\bibitem{adaboost-ids}
W.~{Hu}, W.~{Hu}, and S.~{Maybank}, ``Adaboost-based algorithm for network
  intrusion detection,'' \emph{IEEE Transactions on Systems, Man, and
  Cybernetics, Part B (Cybernetics)}, vol.~38, no.~2, pp. 577--583, April 2008.

\bibitem{haque}
J.~{Zhang}, M.~{Zulkernine}, and A.~{Haque}, ``Random-forests-based network
  intrusion detection systems,'' \emph{IEEE Transactions on Systems, Man, and
  Cybernetics, Part C (Applications and Reviews)}, vol.~38, no.~5, pp.
  649--659, Sep. 2008.

\bibitem{denning}
D.~E. {Denning}, ``An intrusion-detection model,'' \emph{IEEE Transactions on
  Software Engineering}, vol. SE-13, no.~2, pp. 222--232, Feb 1987.

\bibitem{brdiczka}
O.~{Brdiczka}, J.~{Liu}, B.~{Price}, J.~{Shen}, A.~{Patil}, R.~{Chow},
  E.~{Bart}, and N.~{Ducheneaut}, ``Proactive insider threat detection through
  graph learning and psychological context,'' in \emph{2012 IEEE Symposium on
  Security and Privacy Workshops}, May 2012, pp. 142--149.

\bibitem{spitzner}
L.~{Spitzner}, ``Honeypots: catching the insider threat,'' in \emph{19th Annual
  Computer Security Applications Conference, 2003. Proceedings.}, Dec 2003, pp.
  170--179.

\bibitem{granstorm}
K.~{Granström}, P.~{Willett}, and Y.~{Bar-Shalom}, ``Asymmetric threat
  modeling using hmms: Bernoulli filtering and detectability analysis,''
  \emph{IEEE Transactions on Signal Processing}, vol.~64, no.~10, pp.
  2587--2601, May 2016.

\bibitem{zhan}
Z.~{Zhan}, M.~{Xu}, and S.~{Xu}, ``Characterizing honeypot-captured cyber
  attacks: Statistical framework and case study,'' \emph{IEEE Transactions on
  Information Forensics and Security}, vol.~8, no.~11, pp. 1775--1789, Nov
  2013.

\bibitem{kaaniche}
M.~Kaaniche, Y.~Deswarte, E.~Alata, M.~Dacier, and V.~Nicomette, ``Empirical
  analysis and statistical modeling of attack processes based on honeypots,''
  \emph{CoRR}, vol. abs/0704.0861, 01 2007.

\bibitem{thonnard}
O.~Thonnard and M.~Dacier, ``A framework for attack patterns' discovery in
  honeynet data,'' \emph{Digital Investigation}, vol.~5, 09 2008.

\bibitem{almotairi}
S.~{Almotairi}, A.~{Clark}, G.~{Mohay}, and J.~{Zimmermann}, ``A technique for
  detecting new attacks in low-interaction honeypot traffic,'' in \emph{2009
  Fourth International Conference on Internet Monitoring and Protection}, May
  2009, pp. 7--13.

\bibitem{zhang}
Q.~Zhang, D.~Man, and W.~Yang, ``Using hmm for intent recognition in cyber
  security situation awareness,'' \emph{Knowledge Acquisition and Modeling,
  International Symposium on}, vol.~2, pp. 166--169, 11 2009.

\bibitem{ye}
{Nong Ye}, S.~{Vilbert}, and {Qiang Chen}, ``Computer intrusion detection
  through ewma for autocorrelated and uncorrelated data,'' \emph{IEEE
  Transactions on Reliability}, vol.~52, no.~1, pp. 75--82, March 2003.

\bibitem{mitchell}
R.~{Mitchell} and I.~{Chen}, ``Behavior-rule based intrusion detection systems
  for safety critical smart grid applications,'' \emph{IEEE Transactions on
  Smart Grid}, vol.~4, no.~3, pp. 1254--1263, Sep. 2013.

\bibitem{shi}
D.~{Shi}, R.~J. {Elliott}, and T.~{Chen}, ``On finite-state stochastic modeling
  and secure estimation of cyber-physical systems,'' \emph{IEEE Transactions on
  Automatic Control}, vol.~62, no.~1, pp. 65--80, Jan 2017.

\bibitem{jonsson}
E.~{Jonsson} and T.~{Olovsson}, ``A quantitative model of the security
  intrusion process based on attacker behavior,'' \emph{IEEE Transactions on
  Software Engineering}, vol.~23, no.~4, pp. 235--245, April 1997.

\bibitem{qdla}
Q.~D. {La}, T.~Q.~S. {Quek}, J.~{Lee}, S.~{Jin}, and H.~{Zhu}, ``Deceptive
  attack and defense game in honeypot-enabled networks for the internet of
  things,'' \emph{IEEE Internet of Things Journal}, vol.~3, no.~6, pp.
  1025--1035, Dec 2016.

\bibitem{rashid}
\BIBentryALTinterwordspacing
T.~Rashid, I.~Agrafiotis, and J.~R. Nurse, ``A new take on detecting insider
  threats: Exploring the use of hidden markov models,'' in \emph{Proceedings of
  the 8th ACM CCS International Workshop on Managing Insider Security Threats},
  ser. MIST '16.\hskip 1em plus 0.5em minus 0.4em\relax New York, NY, USA: ACM,
  2016, pp. 47--56. [Online]. Available:
  \url{http://doi.acm.org/10.1145/2995959.2995964}
\BIBentrySTDinterwordspacing

\bibitem{krogh}
\BIBentryALTinterwordspacing
A.~Krogh, M.~Brown, I.~Mian, K.~Sjölander, and D.~Haussler, ``Hidden markov
  models in computational biology: Applications to protein modeling,''
  \emph{Journal of Molecular Biology}, vol. 235, no.~5, pp. 1501 -- 1531, 1994.
  [Online]. Available:
  \url{http://www.sciencedirect.com/science/article/pii/S0022283684711041}
\BIBentrySTDinterwordspacing

\bibitem{won}
{Kyoung-Jae Won}, A.~{Prugel-Bennett}, and A.~{Krogh}, ``Evolving the structure
  of hidden markov models,'' \emph{IEEE Transactions on Evolutionary
  Computation}, vol.~10, no.~1, pp. 39--49, Feb 2006.

\bibitem{johansson}
M.~{Johansson} and T.~{Olofsson}, ``Bayesian model selection for markov, hidden
  markov, and multinomial models,'' \emph{IEEE Signal Processing Letters},
  vol.~14, no.~2, pp. 129--132, Feb 2007.

\bibitem{rabiner}
L.~{Rabiner} and B.~{Juang}, ``An introduction to hidden markov models,''
  \emph{IEEE ASSP Magazine}, vol.~3, no.~1, pp. 4--16, Jan 1986.

\bibitem{alghamdi}
R.~Alghamdi, ``Hidden markov models (hmms) and security applications,''
  \emph{International Journal of Advanced Computer Science and Applications},
  vol.~7, 02 2016.

\bibitem{ghahramani}
Z.~Ghahramani, ``An introduction to hidden markov models and bayesian
  networks.'' \emph{IJPRAI}, vol.~15, pp. 9--42, 02 2001.

\bibitem{moon}
T.~K. {Moon}, ``The expectation-maximization algorithm,'' \emph{IEEE Signal
  Processing Magazine}, vol.~13, no.~6, pp. 47--60, Nov 1996.

\bibitem{hinton}
D.~E. Rumelhart, G.~E. Hinton, and R.~J. Williams, ``Learning representations
  by back-propagating errors,'' \emph{Nature}, vol. 323, pp. 533--536, 1986.

\bibitem{cowrie}
\BIBentryALTinterwordspacing
M.~Oosterhof, ``Cowrie - medium-interaction honeypot,'' Jul 2019. [Online].
  Available: \url{https://github.com/micheloosterhof/cowrie}
\BIBentrySTDinterwordspacing

\bibitem{rade}
R.~Rade, S.~Deshmukh, R.~Nene, A.~S. Wadekar, and A.~Unny, ``Temporal and
  stochastic modelling of attacker behaviour,'' in \emph{Advances in Data
  Science}, L.~Akoglu, E.~Ferrara, M.~Deivamani, R.~Baeza-Yates, and P.~Yogesh,
  Eds.\hskip 1em plus 0.5em minus 0.4em\relax Singapore: Springer Singapore,
  2019, pp. 30--45.

\bibitem{shlens}
J.~Shlens, ``A tutorial on principal component analysis,'' \emph{ArXiv}, vol.
  abs/1404.1100, 2014.

\end{thebibliography}


\end{document}